\documentclass[10pt,twocolumn,letterpaper]{article}

\usepackage[pagenumbers]{iccv} 

\definecolor{iccvblue}{rgb}{0.21,0.49,0.74}
\PassOptionsToPackage{table,xcdraw}{xcolor}
\usepackage{xcolor}
\usepackage[pagebackref,breaklinks,colorlinks,allcolors=iccvblue]{hyperref}
\usepackage{multirow}
\usepackage{graphicx}
\usepackage{bbm}
\usepackage{caption}
\usepackage{amsmath}
\usepackage{float}
\usepackage{algorithm}
\usepackage{algpseudocode}

\title{Structure-aware Semantic Discrepancy and Consistency for 3D Medical Image Self-supervised Learning} 

\author{Tan Pan$^{1,2}$,~Zhaorui Tan$^2$,~Kaiyu Guo$^{3,2}$,~Dongli Xu$^2$,~Weidi Xu,\\~Chen Jiang$^{2}$\thanks{\* Corresponding authors. This research was conducted during an internship at the Shanghai Academy of Artificial Intelligence for Science.}, ~Xin Guo$^2$,Yuan Qi$^{1,2,4}$,~Yuan Cheng$^{1,2}$\footnotemark[1]\\
\textsuperscript{1} Artificial Intelligence Innovation and Incubation Institute, Fudan University \\
\textsuperscript{2} Shanghai Academy of Artificial Intelligence for Science \\
\textsuperscript{3} The University of Queensland \\
\textsuperscript{4} Zhongshan Hospital, Fudan University\\
{\tt\small {pant23@m.fudan.edu.cn, jiangchen@sais.com.cn, chengyuan@sais.com.cn}}
}

\begin{document}
\maketitle
\begin{abstract}
3D medical image self-supervised learning (mSSL) holds great promise for medical analysis. Effectively supporting broader applications requires considering anatomical structure variations in location, scale, and morphology, which are crucial for capturing meaningful distinctions. However, previous mSSL methods partition images with fixed-size patches, often ignoring the structure variations. In this work, we introduce a novel perspective on 3D medical images with the goal of learning structure-aware representations. We assume that patches within the same structure share the same semantics (semantic consistency) while those from different structures exhibit distinct semantics (semantic discrepancy). Based on this assumption, we propose an mSSL framework named $S^2DC$, achieving \textbf{S}tructure-aware \textbf{S}emantic \textbf{D}iscrepancy and \textbf{C}onsistency in two steps. First, $S^2DC$ enforces distinct representations for different patches to increase semantic discrepancy by leveraging an optimal transport strategy. Second, $S^2DC$ advances semantic consistency at the structural level based on neighborhood similarity distribution. By bridging patch-level and structure-level representations, $S^2DC$ achieves structure-aware representations. Thoroughly evaluated across 10 datasets, 4 tasks, and 3 modalities, our proposed method consistently outperforms the state-of-the-art methods in mSSL.

\end{abstract}    
\section{Introduction}
\label{sec:intro}
Self-supervised learning (SSL) is widely used in medical image analysis to build strong representations that improve performance on various applications and maximize the use of unlabeled data~\cite{wu2024voco,oord2018representation,wang2023swinmm}. These applications frequently focus on different specific anatomical structures (i.e., organ segmentation), each distinguished across anatomy due to location, scale, morphology, and other variances~\cite{d2024totalsegmentator,simpson2019large,chen2023total,gatidis2023autopet}. For example, the liver is a large, well-defined solid organ with relatively homogeneous intensity and distinct anatomical boundaries. In contrast, veins are thin, tubular structures with varying diameters, lower contrast, and complex branching patterns. 

\begin{figure}
  \centering
  \includegraphics[width=\columnwidth]{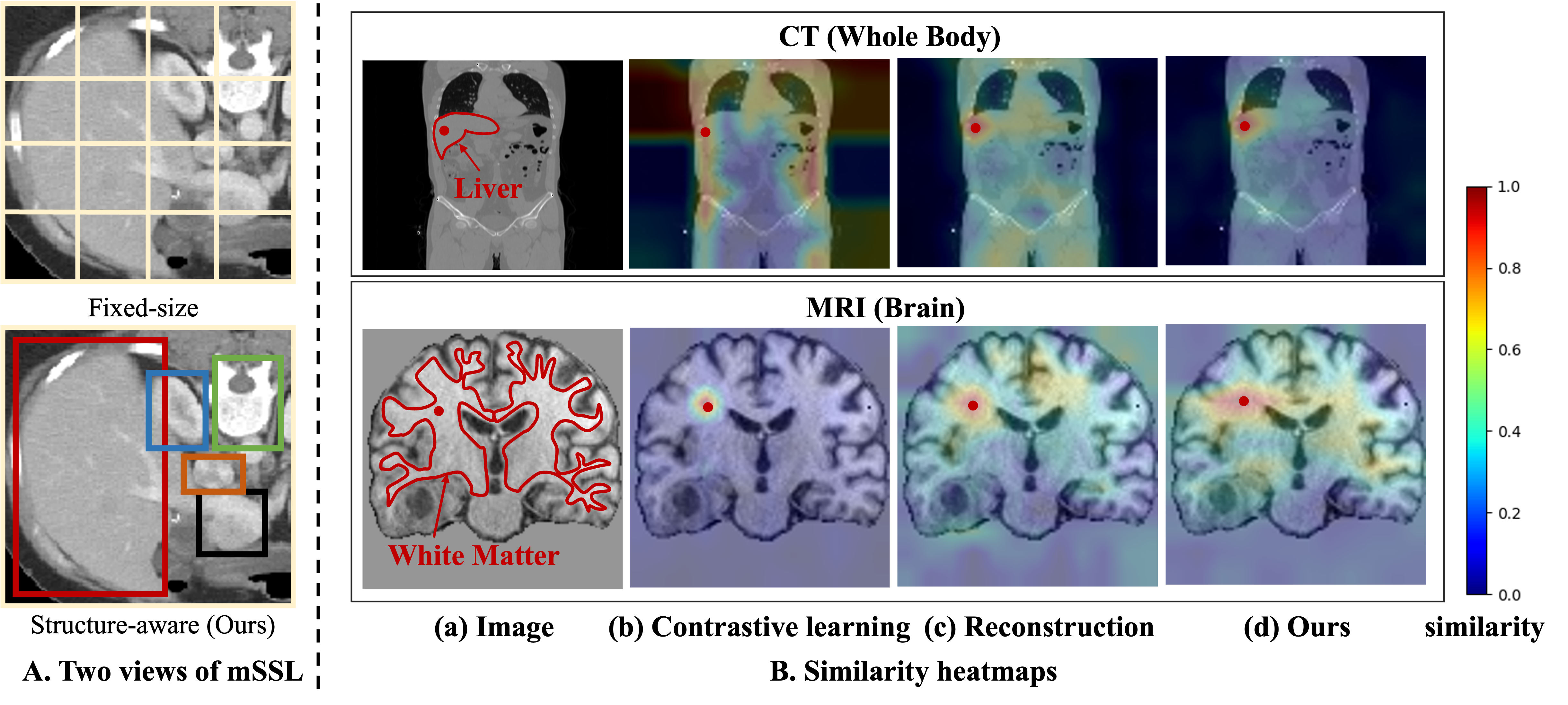}
  \caption{A. Two views of mSSL. B. Similarity heatmaps on CT and MRI images across different methods. 
  We sample an anchor patch (red dots) and compute feature similarities with all other patches. In the first row, the liver anchor should show low similarity with non-liver patches, while in the second row, the white matter anchor should exhibit high similarity with other white matter patches. Current SOTA contrastive-based~\cite{wu2024voco} (b) and reconstruction-based~\cite{wang2023swinmm} (c) methods struggle with both patch feature discrepancy in different structures and consistency in the same structure. In contrast, our method (d) advances both discrepancy and consistency.}
\label{fig:featureconsistency}
\end{figure}




Current medical image SSL (mSSL) methods can be classified into contrastive-based and reconstruction-based methods~\cite{wang2023swinmm,he2023swinunetr,jiang2023anatomical,wu2024voco}. For example,~\citeauthor{wang2023swinmm} utilize multi-view reconstruction to encourage dense representations, while~\citeauthor{wu2024voco} incorporate overlapped patch ratio as soft contrastive constraint to improve semantic consistency. However, these methods use fixed-size image partition, treating all segmented patches uniformly across structures and overlooking their underlying semantic relationship among structures as shown in ~\cref{fig:featureconsistency} A. To illustrate these limitations, we visualize similarity heatmaps on CT and MRI images across different methods by overlaying similarity scores onto the original images (\cref{fig:featureconsistency} B). Ideally, patches within the same organ should exhibit high similarity, while those from different organs should show low similarity. However, as shown in \cref{fig:featureconsistency} (a) and (b), existing methods struggle with maintaining intra-structure patch consistency and distinguishing inter-structure patch differences. A more detailed analysis and metrics of organ-level relationships in the state-of-the-art (SOTA) methods are provided in \cref{fig:tsne}.


Based on these empirical analyses, we propose that patches within the same structure should share the same semantics, while those from different structures should exhibit distinct semantics. To achieve this, patch relationships must be structure-aware and adaptively adjusted to account for anatomical variations. Based on this assumption, we define structure-aware learning through two key principles: intra-structure consistency and inter-structure discrepancy.



To achieve these principles, we propose \textbf{$S^{2}DC$}, a self-supervised learning framework for 3D medical imaging designed to capture \textbf{S}tructure-aware \textbf{S}emantic \textbf{C}onsistency and \textbf{D}iscrepancy. The core idea of our framework is to explore semantic relevance between patches to enhance structure-level representations, which unfolds in two key steps. 
First, we increase the discrepancy between patches with different semantics through patch-to-patch discrepancy. By an optimal transport strategy, patches with different semantics are pushed further apart.
Second, beyond patch-wise distinction, the framework further explores patch-to-structure semantic consistency, leveraging the observation that patches with shared semantics form peaks in the similarity distribution. These distributional characteristics serve as soft regularization, identifying patches belonging to the same structure and enforcing structural-level consistency.  Notably, the principles of inter-structure discrepancy and intra-structure consistency of our framework are also suitable for functional imaging or metabolic imaging.


The contributions of this work can be summarized as follows:
\begin{itemize}
    \item We propose a novel insight of intra-structure consistency and inter-structure discrepancy in the anatomical structure-aware feature learning in 3D medical images. We argue that the patches within the same anatomical structures maintain consistent semantics, while the patches from different anatomical structures reflect distinct semantics.
    \item Based on our insight and assumption, we propose $S^{2}DC$, a novel training framework that enhances inter-structure discrepancy and intra-structure consistency. The framework establishes reliable patch-to-patch correspondences to reinforce discrepancy while leveraging patch-to-structure semantic connectivity from the similarity distribution to improve consistency.
    \item Our method demonstrates superior performance over SOTA medical image SSL methods, evaluated across 10 datasets, 4 tasks, and 3 imaging modalities. The code will be available upon publication.
\end{itemize}


\section{Related work}
\label{sec:formatting}
\textbf{SSL in 3D medical images.} Contrastive-based and reconstruction-based SSL are among the most prominent paradigms in mSSL. Some approaches emphasize the fine-grained capabilities of models, incorporating techniques such as multi-scale reconstruction or reconstruction from disarranged images to enhance model performance \cite{zhuang2024mim,tao2020revisiting,zhuang2023advancing}. Simultaneously, other methods \cite{wu2024voco,zhou2023pcrlv2} prioritize preserving maximal semantic information by leveraging contextual priors. Recently, several studies have explored the combination of contrastive learning (CL) and reconstruction to further improve model performance \cite{wang2023swinmm,jiang2023anatomical,tang2022self,he2023geometric}. These works aim to learn consistent semantic representations by applying contrastive learning to global features or predicting missing information from available context. However, in mSSL, few approaches enforce constraints to enhance both anatomical structure-aware discrepancy and consistency, which is the central focus of this paper.

\textbf{Geometric approaches for medical image analysis.} Our method adopts some geometric characteristics for patch-to-patch correspondence. In radiographic imaging, geometric characteristics, such as location and anatomical structure, play a crucial role in organ identification, disease diagnosis, and medical analysis. Several approaches~\cite{huang2022attentive,santhirasekaram2024geometric} have incorporated consistent and recurrent anatomical patterns as learnable priors to enhance medical image analysis. For example, 
~\citeauthor{huang2022attentive} proposed to enhance segmentation by establishing patch correspondence between two 2D images. Those methods utilizing the geometric or anatomical invariant have shown remarkable improvement in the downstream tasks.  Some medical SSL works~\cite{li2024self,he2023geometric} also explore geometric correspondence for better performance. The SSL method GVSL~\cite{he2023geometric} considers the geometric transformation and reconstructs affined images to original images to learn global and local representations. ~\citeauthor{yang2024keypoint} present a keypoint-augmented fusion layer with the keypoint guidance from SIFT to extract local and global information to improve segmentation results. Those methods leverage the invariance of anatomical structures while not considering the relation between anatomical structures.

\textbf{Multi-scale features in SSL.} Multi-scale features have received much attention in existing SSL research, which is relevant but not sufficient to structure-aware representations. Multi-scale feature learning in SSL can be categorized as reconstruction-based and contrastive-based methods. For instance, the contrastive-based method ~\citeauthor{wang2021dense} introduced a dense constraint based on the correspondence of local features for dense prediction tasks. The method~\cite{wang2024groupcontrast} developed data-driven contrastive learning methods by utilizing prototypes. ~\citeauthor{zhou2023pcrlv2} combines multi-scale reconstruction and multi-scale contrastive learning to learn multi-scale features. Although those methods achieve success in learning multi-scale features, they are not aware of anatomical structure variations because of the fixed-size image partition strategy and uniform contribution of different anatomical structures in images.


\section{Methods} \label{method}
We present our two-step approach: patch-to-patch discrepancy and patch-to-structure consistency. The patch-to-patch constraint increases the distance between all patches, ensuring that patches with different semantics are pushed further apart. In contrast, the patch-to-structure constraint keeps patches within the same structure close together, enforcing semantic consistency within the same structure.
\subsection{Preliminaries}  \label{preliminary}
\textbf{Contrastive Learning.} 
Contrastive learning in SSL constructs positive sample pairs to capture shared information~\cite{tsai2020self,oord2018representation}, typically using a source image and its augmented counterpart as two views of the same image. Based on the theory that the features learned from these paired views suffice for downstream tasks~\cite{tsai2020self}, we focus solely on positive pairs from the same source image rather than different ones. We adopt vanilla InfoNCE in MoCo~\cite{he2020momentum} as our baseline, leveraging a teacher-student mechanism and a queue storing previously generated features. Following prior SSL methods in 3D medical imaging, images are partitioned into multiple sub-volumes to optimize memory efficiency~\cite{wang2023swinmm,wu2024voco,he2023swinunetr}. Given $N$ non-overlapping subvolumes \{$v_k | k=1,2,3...,N$\} cropped from a full volume $\mathcal{V}$, we randomly select a subvolume $v_i$ and generate an augmented version $v_i^{'}$ as a positive pair. The constraint follows a dictionary look-up strategy, where a query feature is matched against candidate features in the dictionary. More details can be found in 
Appendix.





Based on the dictionary $\{q_m^*|m=1,2,...,K\}$ where $q_m^*$ is the feature encoded by the teacher model and $K$ is the size of the dictionary and the query $q_i$ encoded by the student model, the InfoNCE loss is conducted as a contrastive loss $\mathcal{L}_g$:
\begin{equation}
\begin{split}
    \mathcal{L}_g = -\log{\frac{\exp{\left ( q_i\cdot{q_i^{*'}}\right / \tau )}}{\sum_{m=0}^{K}{\exp{\left ( q_i\cdot{q_m^*} \right ) / \tau}}}},
\end{split}
\end{equation}
where $\tau$ is a temperature hyper-parameter \cite{he2020momentum}.

\textbf{Geometric equivariance.} Mao et al. \cite{mao2023robust} propose that the model satisfies equivariance, a meta-property that can be applied to dense feature maps and generalized to most existing vision tasks. Equivariance implies that geometric transformations applied to the input image are preserved in the output features, which benefits many tasks \cite{chaman2021truly,mao2023robust}. Consequently, we expect the token features from the Transformer encoder $\mathcal{\varphi}$ to undergo the same homography transformation as the input images. Based on this assumption, we can establish correspondences between token features in the two input volumes to compute patch-to-patch and patch-to-structure relationships. Remarkably, we consider each token to correspond to a patch in the input volume.

\begin{figure*}[!htbp]
  \centering
  \includegraphics[width=0.85\textwidth]{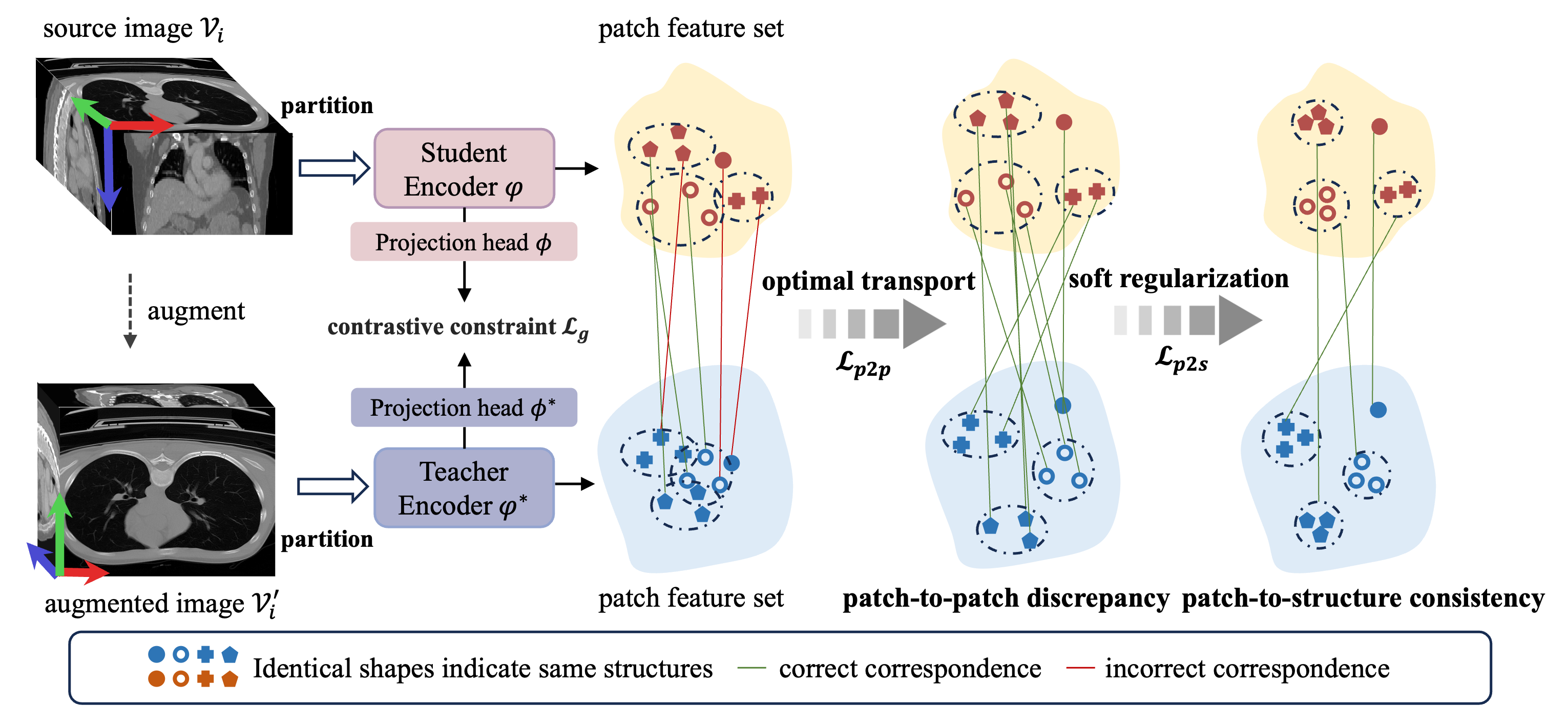}
  \vskip -0.05in
  \caption{The pipeline of our SSL framework. $S^{2}DC$ is established on patch features (i.e., token feature in vision transformer) and incorporates two main steps: (1) Patch-to-patch discrepancy. The model aligns two patch sets, which treats every patch as distinct semantics to enlarge the distribution gap between all semantics. (2) Patch-to-structure consistency. The model applies a soft regularization to align patches within the same structure. By those two steps, the framework captures semantic connectivity between anatomical structures.}
  \label{fig:framework}
  \vskip -0.1in
\end{figure*}

\subsection{Patch-to-patch discrepancy} \label{onetoone}
Given a volume $\mathcal{V}_i$ and its augmented counterpart $\mathcal{V}_i^{'}$, we build patch-to-patch correspondence between them. A reliable patch-to-patch correspondence refers to correct one-to-one alignment among two patch sets, which treats every patch as a distinct object. This step aims to enlarge the distribution gap between all semantics and capture distinct characteristics. 

In mSSL, augmentations can be broadly classified into geometric and intensity transformations, with only geometric transformations affecting voxel positions. Geometric transformations involve modeling an affine matrix $H$, which is applied to the target image to alter its spatial configuration. Therefore, in this section, we discuss the situation after applying an affine transformation. 

\textbf{Ground truth of patch correspondence.}  Given two patch centers $c_i=(x_i,y_i,z_i)$ and $c_j=(x_j,y_j,z_j)$ from $\mathcal{V}_i$ and $\mathcal{V}_i^{'}$, where $x$, $y$, and $z$ are coordinates, the correspondence between two patches is determined based on the re-projection distance of their central positions mapped in the input volume, as described in  \cref{mnn_correspondence} and \cref{getcoord}.


\begin{equation}
    \mathcal{M}_{gt}{(i,j)}\triangleq\left\{
    \begin{aligned}
    &1 && if \left \langle H(c_i),c_j\right \rangle  \wedge  \left \langle H^{-1}(c_j),c_i\right \rangle,  \\
    &0 && else.
    \end{aligned} \label{mnn_correspondence}
    \right.
\end{equation}
Here, $H^{-1}(\cdot)$ denotes the inverse of the affine transformation $H(\cdot)$ in 3D space, yielding the transformed position. The notation $\left \langle \cdot,\cdot \right \rangle$ serves as the metric to determine whether two patches originate from the same patch. Specifically, the metric is:
\begin{equation}
\left \langle H(c_i),c_j\right \rangle \!\!\triangleq \!\!\left\{
    \begin{aligned}
    &1,if (\bigtriangleup{x_{ij}},\!\bigtriangleup{y_{ij}},\! \bigtriangleup{z_{ij}}) \\
    &
\!\in\! [-\frac{W}{2},\!\frac{W}{2}]\!\times\! [-\frac{H}{2},\!\frac{H}{2}]\!\times\![-\frac{D}{2},\frac{D}{2}].\\
    &0, else,
    \end{aligned} \label{getcoord}
    \right.
\end{equation}
Here, $W$, $H$ and $D$ are the width, height and depth of patch and $\bigtriangleup{x_{ij}}, \bigtriangleup{y_{ij}}$ and $\bigtriangleup{z_{ij}}$ are the coordinate differences of $H(c_i)$ and $c_j$. With \cref{mnn_correspondence} and \cref{getcoord}, the ground-truth patch-to-patch correspondences $\mathcal{M}_{gt}$ between two images are obtained.

\textbf{Building reliable patch-to-patch correspondences.} Given token features from $\varphi^*(\mathcal{V}_i^{'})$ and $\varphi(\mathcal{V}_i)$, we can calculate the similarity map $\mathcal{M}_t$ between tokens and get the loss between $\mathcal{M}_t$ and $\mathcal{M}_{gt}$. However, similar to the issues in image matching~\cite{sun2021loftr,sarlin2020superglue}, reliably identifying patch correspondence in medical images is challenging due to repetitive patterns. This intractable challenge would be optimized by solving a differentiable optimal transport problem, such as the Sinkhorn algorithm~\cite{sinkhorn1964relationship}. The Sinkhorn algorithm is used to solve entropy-regularized optimal transport, enabling soft assignment of correspondences between feature points by iteratively refining a doubly stochastic matrix, improving robustness to noise and deformations. The efficiency of Sinkhorn is affected by the large number of iterations.

\begin{figure*}[!htbp]
  \centering
  \includegraphics[width=0.85\textwidth]{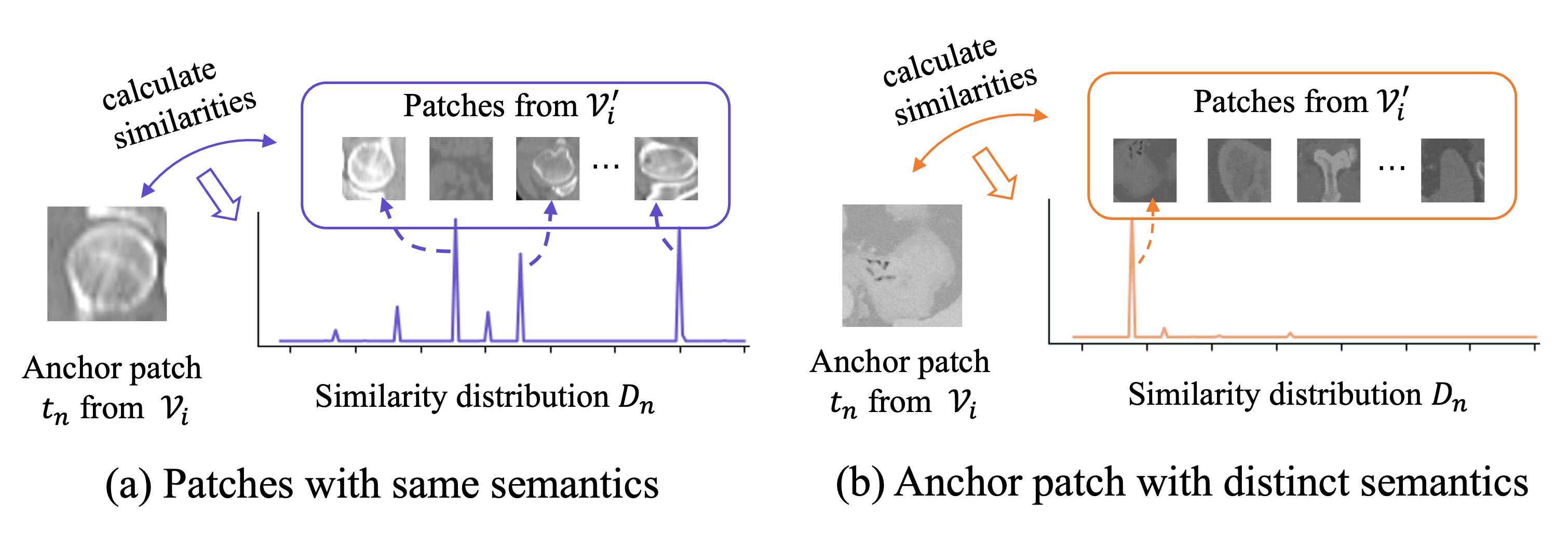}
  \vskip -0.1in
  \caption{Illustration of the similarity distribution $D_n$. (a) Patches with the same semantics (e.g., bone). Given an anchor patch, patches from the same semantics form several peaks in $D_n$. (b) Anchor patch with distinct semantics (e.g., pancreas), $D_n$ shows only one large peak (its augmented patch). }
  \label{fig:similarity_distribution}
  \vskip -0.15in
\end{figure*}

~\citeauthor{rocco2018neighbourhood} propose using a dual-softmax operator to establish reliable correspondences based on neighborhood consensus, which brings mutual nearest neighbors closer to learn a more compact similarity distribution. Given the similarity map $\mathcal{M}_t$ between two images, the similarities after applying the dual-softmax operator is:\\
\begin{equation}\label{mnn}
\mathcal{\hat{M}}_t(i,j)=softmax(\mathcal{M}_t\!(i,\!\cdot\!))\cdot\!softmax((\mathcal{M}_t(\!\cdot\!,j)).
\end{equation}

Based on the ground-truth correspondences ${\mathcal{M}_{gt}}$ and the re-assigned feature similarity matrix $\mathcal{\hat{M}}_t$, we use the cross-entropy loss to minimize Kullback-Leibler divergence (KL divergence) between them~\cite{thomas2006elements}. This patch-to-patch loss $\mathcal{L}_{p2p}$ is:
\begin{equation}\label{dense_constraint}
    \mathcal{L}_{p2p} = -\frac{1}{|\mathcal{M}_{gt}|} \sum_{i=0}^N\sum_{j=0}^N{\mathcal{M}_{gt}(i,j)\times log(\mathcal{\hat{M}}_t(i,j))},
\end{equation}
where $N$ represents the number of patches (i.e., tokens in the 3D Transformer).

\subsection{Patch-to-structure consistency}\label{one2all}
The correspondence assignment focuses on enhancing reliable patch-to-patch correspondence, which treats every patch as distinct. Patch-to-patch discrepancy limits the exploration of connectivity among patches from the same semantics (\textit{i.e.}, same anatomical structure). Thus, a patch-to-structure correspondence should be explored to improve consistency.

We define neighborhood similarity distribution as $\mathcal{D}_n$, which represents the similarity vector of an anchor patch feature $t_n$ from $\mathcal{V}_i$ with all the patch features from $\mathcal{V}_i^{'}$. As shown in \cref{fig:similarity_distribution}, patch pairs with the same semantics form distinct peaks in the distribution $\mathcal{D}_n$, which meets the assumption that the same structures have the same semantics. Thus, the central tendency and dispersion of $\mathcal{D}_n$ can be employed to constrain semantic connectivity. For instance, if $t_n$ is similar to several patches of $\mathcal{V}_i^{'}$, indicating they may belong to the same anatomical structure, $\mathcal{D}_n$ tends to be more dispersed (~\cref{fig:similarity_distribution} (b)). In contrast, with no semantic-similar patches, the distribution of $t_n$ is more concentrated. Therefore, we utilize the characteristics of $\mathcal{D}_n$ as a soft label to regularize the patch-to-patch loss $\mathcal{L}_{p2p}$ to maintain underlying semantic relevance.

We introduce a simplified version of the Sharpe ratio~\cite{sharpe1966mutual} to reflect the dispersion and concentration of neighborhood similarity distribution $\mathcal{D}_n$. Given the $\mathcal{D}_n$ of the patch feature $t_n$, the Sharpe ratio of $t_n$ can be computed as:
\begin{equation}\label{Sharpe}
    sr_{\mathcal{V}_i}^n=\frac{\max(\mathcal{D}_n)-\frac{1}{N}\sum_{m=1}^{N}\mathcal{D}_n}{\sigma_{\mathcal{D}_n}},
\end{equation}
where, $\sigma_{\mathcal{D}_n}$ is the variance of vector $\mathcal{D}_n$ and the notation $\max(\mathcal{D}_n)$ is the maximal value of $\mathcal{D}_n$. A low Sharpe ratio of $\mathcal{D}_n$ indicates that the distribution is more dispersed, which means the patch feature $t_i$ has more than one similar semantic patch feature. More details can be seen in Appendix.

$sr_{\mathcal{V}_i}^n$ is utilized as the soft regularization to adaptively adjust 
$\mathcal{L}_{p2p}$, serving as soft evidence for maintaining a semantic connectivity relationship among patch features. The weighted losses for the feature $t_n$ of the volume $\mathcal{V}_i$ is $\mathcal{L}_c(n, \cdot) = softmax(sr_{\mathcal{V}_i})_n\mathcal{L}_{p2p}(n,\cdot)$, where $sr_{\mathcal{V}_i} = \{sr_{\mathcal{V}_i}^n\}_n$. Similarly, we have $\mathcal{L}_c(\cdot, m) = softmax(sr_{\mathcal{V}_i^{'}})_m\mathcal{L}_{p2p}(\cdot, m)$. Thus, we have the loss of a single pair:

\begin{equation}
    l_{n,\! m}\!=\frac{\! (softmax(sr_{\mathcal{V}_i})_n\!+\!softmax(sr_{\mathcal{V}_i^{'}})_m)\mathcal{L}_{p2p}(n,m)}{2}.
\end{equation}
Finally, the patch-to-structure loss\ $\mathcal{L}_{p2s}$ is

\begin{equation}
\begin{split}
    \mathcal{L}_{p2s} = \frac{\sum_{m=0}^N\sum_{n=0}^N{l_{n,m}}}{N\times{N}}.
\end{split}
\end{equation}

\subsection{Overall loss function} \label{overallloss} 
The final combined objective for the proposed framework is defined as:
\begin{equation}
    \mathcal{L} = \mathcal{L}_g+\mathcal{L}_{p2p} +\mathcal{L}_{p2s},
\end{equation}

\section{Experiments}
\begin{table*}[!ht]
\centering
\resizebox{\textwidth}{!}{%
\begin{tabular}{lcccccccccccccc}
\hline
 & \multicolumn{14}{c}{\textbf{Dice Score(\%)}} \\ \cline{2-15} 
\multirow{-2}{*}{\textbf{Method}} & Spl & RKid & LKid & Gall & Eso & Liv & Sto & Aor & IVC & Veins & Pan & RAG & LAG & \textbf{AVG} \\ \hline
\multicolumn{1}{c}{\textbf{}} & \multicolumn{14}{c}{\textbf{From Scratch}} \\
UNETR~\cite{hatamizadeh2022unetr} & 93.02 & 94.13 & 94.12 & 66.99 & 70.87 & 96.11 & 77.27 & 89.22 & 82.1 & 70.16 & 76.65 & 65.32 & 59.21 & 79.82 \\
Swin-UNETR~\cite{he2023swinunetr} & 94.06 & 93.54 & 93.8 & 65.61 & 74.6 & 97.09 & 75.94 & 91.8 & 82.36 & 73.63 & 75.19 & 68 & 61.11 & 80.53 \\ \hline
\multicolumn{1}{c}{\textbf{}} & \multicolumn{14}{c}{\textbf{With General SSL}} \\
MAE3D~\cite{chen2023masked} & 93.98 & 94.37 & 94.18 & 69.86 & 74.64 & 96.66 & 80.4 & 90.3 & 83.13 & 72.65 & 77.11 & 67.34 & 60.54 & 81.33 \\
SimCLR~\cite{chen2020simple} & 92.79 & 93.04 & 91.41 & 49.65 & 50.99 & 98.49 & 77.92 & 85.56 & 80.58 & 64.37 & 67.16 & 59.04 & 48.99 & 73.85 \\
SimMIM~\cite{xie2022simmim} & 95.56 & 95.82 & 94.14 & 52.06 & 53.52 & 98.98 & 80.25 & 88.11 & 82.98 & 66.49 & 69.16 & 60.88 & 50.45 & 76.03 \\
MoCo v3~\cite{he2020momentum} & 91.96 & 92.85 & 92.42 & 68.25 & 72.77 & 94.91 & 78.82 & 88.21 & 81.59 & 71.15 & 75.76 & 66.48 & 58.81 & 79.54 \\
Jigsaw~\cite{chen2021jigsaw} & 94.62 & 93.41 & 93.55 & 75.63 & 73.21 & 95.71 & 80.8 & 89.41 & 84.78 & 71.02 & 79.57 & 65.68 & 60.22 & 81.35 \\
PositionLabel~\cite{zhang2023positional} & 94.35 & 93.15 & 93.21 & 75.39 & 73.24 & 95.76 & 80.69 & 88.8 & 84.04 & 71.18 & 79.02 & 65.11 & 60.07 & 81.09 \\ \hline
\multicolumn{1}{c}{\textbf{}} & \multicolumn{14}{c}{\textbf{With medical image SSL}} \\
MG~\cite{zhou2021models} & 91.99 & 93.52 & 91.81 & 65.11 & 76.41 & 95.98 & {\color[HTML]{9A0000} \textbf{86.88}} & 89.29 & 83.59 & 71.79 & 81.62 & 67.97 & 63.18 & 81.45 \\
ROT~\cite{taleb20203d} & 91.75 & 93.13 & 91.62 & 65.09 & \underline{76.55} & 94.21 & \underline{86.16} & 89.74 & 83.08 & 71.13 & 81.55 & 67.9 & 63.72 & 81.20 \\
Vicregl~\cite{bardes2022vicregl} & 90.32 & 94.15 & 91.3 & 65.12 & 75.41 & 94.76 & 86.00 & 89.13 & 82.54 & 71.26 & 81.01 & 67.66 & 63.08 & 80.89 \\
Rubik++~\cite{tao2020revisiting} & \underline{96.21} & 90.41 & 89.33 & 75.22 & 72.64 & {\color[HTML]{9A0000} \textbf{97.44}} & 79.25 & 89.65 & 83.76 & 74.74 & 78.35 & 67.14 & 61.97 & 81.38 \\
PCRLv1~\cite{zhou2021preservational} & 95.73 & 89.66 & 88.53 & \underline{75.41} & 72.33 & 96.20 & 78.99 & 89.11 & 83.06 & 74.47 & 77.88 & 67.02 & 61.85 & 80.78 \\
PCRLv2~\cite{zhou2023pcrlv2} & 95.50 & 91.43 & 89.52 & 76.15 & 73.54 & 97.28 & 79.64 & 90.16 & 84.17 & 75.20 & 78.71 & 68.74 & 62.93 & 81.74 \\
Swin-UNETR~\cite{he2023swinunetr} & 95.25 & 93.16 & 92.97 & 63.62 & 73.96 & 96.21 & 79.32 & 89.98 & 83.19 & \underline{76.11} & \underline{82.25} & 68.99 & 65.11 & 81.54 \\
SwinMM~\cite{wang2023swinmm} & 94.33 & 94.18 & 94.16 & 72.97 & 74.75 & 96.37 & 83.23 & 89.56 & 82.91 & 70.65 & 75.52 & 69.17 & 62.90 & 81.81 \\
GL-MAE~\cite{zhuang2023advancing} & 94.54 & 94.39 & 94.37 & 73.19 & 74.93 & 96.51 & 83.49 & 89.74 & 83.11 & 70.80 & 75.71 & \color[HTML]{9A0000} \textbf{69.39} & 63.12 & 82.01 \\
GVSL~\cite{he2023geometric} & 95.27 & 91.22 & 92.25 & 72.69 & 73.56 & 96.44 & 82.40 & 88.90 & 84.22 & 70.84 & 76.42 & 67.48 & 63.25 & 81.87 \\
VoCo~\cite{wu2024voco}& 95.73 & {\color[HTML]{9A0000} \textbf{96.53}} & \underline{94.48} & {\color[HTML]{9A0000} \textbf{76.02}} & 75.60 & \underline{97.41} & 78.43 & {\color[HTML]{9A0000} \textbf{91.21}} & \underline{86.12} & {\color[HTML]{9A0000} \textbf{78.19}} & 80.88 & \underline{71.47} & \underline{67.88} & \underline{83.85} \\
\textbf{$S^{2}DC$} & {\color[HTML]{9A0000} \textbf{96.24}} & \underline{96.51} & {\color[HTML]{9A0000} \textbf{94.52}} & 74.27 & {\color[HTML]{9A0000} \textbf{77.69}} & 96.95 & 83.33 & \underline{90.44} & {\color[HTML]{9A0000} \textbf{86.66}} & 75.22 & {\color[HTML]{9A0000} \textbf{84.24}} & 68.22 & {\color[HTML]{9A0000} \textbf{69.73}} & {\color[HTML]{9A0000} \textbf{84.14}} \\ \hline
\end{tabular}}
\caption{Results on the segmentation task BTCV fine-tuning from pre-training models trained on 1k CT data. Most of the results are drawn from~\cite{zhang2023dive,wu2024voco}. \textcolor[HTML]{9A0000}{\textbf{Best}} and \underline{second best} are highlighted}\label{BTCV} 
\end{table*}
This section begins with an overview of the datasets utilized for both pre-training and downstream tasks. Next, we provide a brief summary of the implementation details for the method. Lastly, we present a comprehensive comparison of the experimental results for our proposed method against other SOTA SSL approaches in 3D medical imaging.

\begin{table}[!htbp]
\centering
\resizebox{0.75\columnwidth}{!}{%
\begin{tabular}{lcc}
\hline
\textbf{Task} & \textbf{Dataset} & \textbf{Modality} \\ \hline
\multirow{6}{*}{Segmentation} & BTCV~\cite{landman2015miccai} & CT \\
 & MSD-Liver~\cite{simpson2019large} & CT \\
 & MSD-Lung~\cite{simpson2019large} & CT \\
 & MSD-Spleen~\cite{simpson2019large} & CT \\
 & BraTs 21~\cite{simpson2019large} & MRI \\
 & AUTOPET~\cite{gatidis2023autopet} & PET \\ \hline
\multirow{1}{*}{Classification} & CC-CCII~\cite{zhang2020clinically} & CT \\
 & ADNI-cls~\cite{mueller2005alzheimer} & PET \\ \hline
Reconstruction & UDPET~\cite{chen2023total} & PET \\ \hline
I2I translation & BraTs 23~\cite{simpson2019large} & MRI \\ \hline
\end{tabular}%
}
\caption{The downstream tasks and modalities. I2I translation represents image-to-image translation.}\label{downstream_datasets}
\vskip -0.15in
\end{table}

\subsection{Datasets and implementation details}
\textbf{Pre-training and downstream datasets.}
Although we emphasize the anatomical structure-specific semantic connectivity, the training process of our mining semantic connective strategy is also suitable for positron emission tomography (PET). In our experiments, we include computed tomography (CT), magnetic resonance imaging (MRI), and PET modalities.

We use the same dataset settings as VoCo~\cite{wu2024voco} on CT, \textit{i.e.}, 10k data collection. BTCV and TCIA COVID-19 are used as pre-training datasets (1k data) for the BTCV segmentation task, while all other CT segmentation tasks are pre-trained on 10k data. Additionally, we collect 6605 (6k) unlabeled MRI scans from OASIS3~\cite{lamontagne2019oasis} and 3519 (3k) PET scans from UDPET~\cite{chen2023total} and ADNI~\cite{mueller2005alzheimer} for pre-training. We evaluate methods on 10 datasets in 3 modalities, as shown in \cref{downstream_datasets}. More details about dataset splits are provided in Appendix.

\textbf{Training setup.} For pre-training stage, $S^{2}DC$ adopts Swin-B~\cite{hatamizadeh2021swin} as the default backbones in three modalities, following previous works~\cite{wu2024voco,hatamizadeh2021swin,wang2023swinmm}.  We set 100k iterations for pre-training in all modalities and crop subvolumes with size $96\times96\times96$. In our patch-to-patch loss $\mathcal{L}_{p2p}$, we adopt the dual-softmax operator for efficiency, while the results of the Sinkhorn algorithm can be found in Appendix. For downstream tasks, we adopt Swin-B and its derivative architectures. The weights of models are initialized by the weights of pre-training models. We adopt the Swin-UNETR~\cite{he2023swinunetr} framework in segmentation, image-to-image translation, and reconstruction tasks. Swin-B~\cite{hatamizadeh2021swin} is used as the backbone of classification tasks. More details about pre-training and training settings can be found in Appendix.

\begin{table*}[!ht]
\centering
\resizebox{0.9\textwidth}{!}{%
\begin{tabular}{lccccccccc}
\hline
 & \multicolumn{9}{c}{\textbf{Dice(\%)}} \\ \cline{2-10} 
 &  &  & \multicolumn{1}{c|}{} & \multicolumn{1}{c|}{} & \multicolumn{4}{c|}{\textbf{BraTs21}} & \textbf{Total} \\ \cline{6-10} 
\multirow{-3}{*}{\textbf{Method}} & \multirow{-2}{*}{\textbf{MSD Liver}} & \multirow{-2}{*}{\textbf{MSD Lung}} & \multicolumn{1}{c|}{\multirow{-2}{*}{\textbf{MSD Spleen}}} & \multicolumn{1}{c|}{\multirow{-2}{*}{\textbf{AUTOPET}}} & \textbf{WT} & \textbf{TC} & \textbf{ET} & \multicolumn{1}{c|}{\textbf{AVG}} & \textbf{AVG} \\ \hline
 & \multicolumn{9}{c}{\textbf{From scratch}} \\
Swin-UNETR~\cite{he2023swinunetr} & 81.18 & 59.73 & \multicolumn{1}{c|}{94.20} & \multicolumn{1}{c|}{34.49} & \color[HTML]{9A0000} \textbf{92.89} & 88.28 & 77.34 & \multicolumn{1}{c|}{86.17} & 71.15 \\ \hline
 & \multicolumn{9}{c}{\textbf{With 3D medical image SSL}} \\
Swin-UNETR~\cite{he2023swinunetr} & 81.58 & 62.46 & \multicolumn{1}{c|}{94.49} & \multicolumn{1}{c|}{43.85} & 92.70 & 88.42 & 78.82 & \multicolumn{1}{c|}{86.64} & 73.95 \\
Swin-MM~\cite{wang2023swinmm} & 82.32 & 61.16 & \multicolumn{1}{c|}{94.60} & \multicolumn{1}{c|}{44.37} & \color[HTML]{9A0000} \textbf{92.89} & \underline{88.54} & \underline{79.18} & \multicolumn{1}{c|}{\underline{86.87}} & 73.86 \\
PCRLv2~\cite{zhou2023pcrlv2} & 82.12 & 62.39 & \multicolumn{1}{c|}{95.23} & \multicolumn{1}{c|}{42.64} & 92.81 & 88.49 & 78.74 & \multicolumn{1}{c|}{86.68} & 73.81 \\
VoCo~\cite{wu2024voco} & \underline{82.50} & \underline{63.34} & \multicolumn{1}{c|}{\underline{95.70}} & \multicolumn{1}{c|}{\underline{45.10}} & 92.90 & 88.44 & 78.62 & \multicolumn{1}{c|}{86.65} & \underline{74.66} \\
\textbf{$S^{2}DC$} & {\color[HTML]{9A0000} \textbf{83.43}} & {\color[HTML]{9A0000} \textbf{64.40}} & \multicolumn{1}{c|}{{\color[HTML]{9A0000} \textbf{95.73}}} & \multicolumn{1}{c|}{{\color[HTML]{9A0000} \textbf{46.47}}} & 92.76 & {\color[HTML]{9A0000} \textbf{88.71}} & {\color[HTML]{9A0000} \textbf{79.79}} & \multicolumn{1}{c|}{\color[HTML]{9A0000} \textbf{87.09}} & \color[HTML]{9A0000} \textbf{75.47} \\ \hline
\end{tabular}%
}
\caption{Results on the segmentation tasks MSD-Liver, MSD-Lung, MSD-Spleen, AUTOPET, and BraTs21. WT, TC, and ET represent the whole tumor, tumor core, and enhancing tumor, respectively. We retrained all methods using 10k, 6k, and 3k data in three modalities separately, except for VoCo in CT, whose weight is provided in the official code. \textcolor[HTML]{9A0000}{\textbf{Best}} and \underline{second best} are highlighted.} \label{segmentation_others}
\vskip -0.15in
\end{table*}

\textbf{Comparison methods.} We compare $S^{2}DC$ with SOTA SSL methods in general~\cite{chen2023masked,chen2020simple,xie2022simmim,he2020momentum,chen2021jigsaw,zhang2023positional} and medical~\cite{zhou2021models,taleb20203d,bardes2022vicregl,tao2020revisiting,zhou2021preservational,zhou2023pcrlv2,he2023swinunetr,wang2023swinmm,zhuang2023advancing,he2023geometric,wu2024voco} images. Meanwhile, we report results of training from scratch (i.e., without initialized weights of pre-training models).

\subsection{Results on downstream tasks.}
We report the results of 4 downstream tasks on 3 modalities. Following previous work, we use the metrics Dice score for segmentation, accuracy for classification, PSNR, SSIM, and NMSE for I2I translation, and SSIM and PSNR for reconstruction. 

\textbf{Segmentation tasks.} Compared to volume-level and voxel-level methods, $S^{2}DC$ performs well both in small targets (AUTOPET, MSD-Lung) and large organs (BTCV, MSD-Liver, MSD-Spleen and BraTs 21) segmentation, as shown in ~\cref{BTCV} and~\cref{segmentation_others}. In BTCV datasets, compared to general SSL methods, medical SSL methods show superior performance. Notably, with $S^{2}DC$ pre-training, we achieve an average improvement of 4.32\%, reaching a Dice score of 75.47\%, which surpasses other SOTA medical SSL methods by 0.81\%.

\textbf{Classification tasks.} We conduct experiments on the CC-CCII dataset for COVID-19 classification and ADNI for AD (Alzheimer's Disease), MCI (Mild Cognitive Impairment), and CN (Cognitively Normal) classification. With the pre-trained model, $S^{2}DC$ achieves an average improvement of 3.93\% and outperforms other medical SSL methods. 
\begin{table}[H]
\centering
\resizebox{0.95\columnwidth}{!}{%
\begin{tabular}{lcccc}
\hline
 &  & \textbf{CC-CCII} & \textbf{ADNI-cls} & \textbf{AVG} \\ \cline{3-5} 
\multirow{-2}{*}{\textbf{Method}} & \multirow{-2}{*}{\textbf{Network}} & \textbf{Acc(\%)} & \textbf{Acc(\%)} & \textbf{Acc(\%)} \\ \hline
 & \multicolumn{3}{c}{\textbf{From Scratch}} \\
Swin-UNETR~\cite{he2023swinunetr} & Swin-B & 88.04 &  56.17 & 72.10  \\ \hline
\textbf{} & \multicolumn{3}{c}{\textbf{With 3D medical image SSL}} \\
 PCRLv2~\cite{zhou2023pcrlv2} & Swin-B & 93.07 &  56.05  & 74.56 \\
Swin-UNETR~\cite{he2023swinunetr} & Swin-B & 94.15 & 55.95  & 75.05  \\
SwinMM~\cite{wang2023swinmm} & Swin-B & \underline{94.80} & \underline{56.36}  & \underline{75.58}  \\
VoCo~\cite{wu2024voco} & Swin-B & 94.60 & 56.32  & 75.46  \\
\textbf{$S^{2}DC$} & Swin-B & {\color[HTML]{9A0000} \textbf{95.34}} & \color[HTML]{9A0000} \textbf{56.72 } & \color[HTML]{9A0000} \textbf{ 76.03 } \\ \hline
\end{tabular}%
}
\vskip -0.05in
\caption{Experimental results on the classification task CC-CCII and ADNI-cls fine-tuning from pre-training models trained on 6k MRI scans. \textcolor[HTML]{9A0000}{\textbf{Best}} and \underline{second best} are highlighted.} \label{CC-CCII}
\vskip -0.05in
\end{table}

\begin{table}[!htbp]
\centering
\resizebox{\columnwidth}{!}{%
\begin{tabular}{lcccccc}
\hline
 & \multicolumn{3}{c}{\textbf{T1-\textgreater{}T2}} & \multicolumn{3}{c}{\textbf{T1-\textgreater{}T1ce}} \\
\multirow{-2}{*}{\textbf{Method}} & PSNR$\uparrow$ & NMSE$\downarrow$ & SSIM$\uparrow$ & PSNR$\uparrow$ & NMSE$\downarrow$ & SSIM$\uparrow$ \\ \cline{2-7} 
 & \multicolumn{6}{c}{\textbf{From Scratch}} \\
Swin-UNETR~\cite{he2023swinunetr} & \underline{26.57} & \underline{0.1348} & 0.8807 & 26.91 & 0.082 & 0.8871 \\ \hline
 & \multicolumn{6}{c}{\textbf{With 3D medical image SSL}} \\
PCRLv2~\cite{zhou2023pcrlv2} & \multicolumn{1}{c}{\color[HTML]{9A0000} \textbf{26.79}} & \multicolumn{1}{c}{0.1640} & \multicolumn{1}{c}{\underline{0.8828}} & \multicolumn{1}{c}{29.62} & \multicolumn{1}{c}{0.123} & \multicolumn{1}{c}{0.9179} \\
Swin-UNETR~\cite{he2023swinunetr}  & 24.81 & 0.2938 & 0.8441  & 30.42 & 0.108 & 0.9117 \\
SwinMM~\cite{wang2023swinmm} & 24.05 & 0.3158 & 0.8182 & 30.70 & 0.099 & 0.9059 \\
VoCo~\cite{wu2024voco} & 25.81 & 0.1916 & 0.8598 & \underline{31.92} & \underline{0.079} & \underline{0.9315} \\
\textbf{$S^{2}DC$} & 25.96 & {\color[HTML]{9A0000} \textbf{0.1161}} & {\color[HTML]{9A0000} \textbf{0.8839}} & {\color[HTML]{9A0000} \textbf{32.65}} & {\color[HTML]{9A0000} \textbf{0.041}} & {\color[HTML]{9A0000} \textbf{0.9412}} \\ \hline
\end{tabular}%
}
\vskip -0.05in
\caption{Experimental results on the image-to-image translation task BraTs23 fine-tuning from pre-training models trained on 6k MRI scans. \textcolor[HTML]{9A0000}{\textbf{Best}} and \underline{second best} are highlighted.} \label{BraTs23}
\vskip -0.1in
\end{table}
\begin{table}[!htbp]
\centering
\resizebox{0.85\columnwidth}{!}{%
\begin{tabular}{lccc}
\hline
\textbf{Method} & \textbf{Network} & \textbf{SSIM$\uparrow$} & \textbf{PSNR$\uparrow$} \\ \hline
\multicolumn{4}{c}{\textbf{From Scratch}} \\
Swin-UNETR~\cite{he2023swinunetr} & - & 0.99917 & \underline{58.4100} \\ \hline
\multicolumn{4}{c}{\textbf{With 3D medical image SSL}} \\
PCRLv2~\cite{zhou2023pcrlv2} & Swin-UNETR & 0.99916 & 58.2865 \\
Swin-UNETR~\cite{he2023swinunetr} & Swin-UNETR & 0.99917 & 58.3409 \\
SwinMM~\cite{wang2023swinmm} & Swin-UNETR & 0.99917 & 58.4002 \\
VoCo~\cite{wu2024voco} & Swin-UNETR & 0.99916 & 58.2561 \\
\textbf{$S^{2}DC$} & Swin-UNETR & \color[HTML]{9A0000} \textbf{0.99917} & {\color[HTML]{9A0000} \textbf{58.4649}} \\ \hline
\end{tabular}%
}
\vskip -0.05in
\caption{Experimental results on low-dose to full-dose reconstruction task UDPET fine-tuning from pre-training models trained on 3k PET scans. \textcolor[HTML]{9A0000}{\textbf{Best}} and \underline{second best} are highlighted.} \label{UDPET}
\vskip -0.15in
\end{table}

\textbf{Image-to-image translation task.} We evaluate in BraTs23, which includes four structural MRI modalities for each individual. The translation quality for T1 $\to$ T2 and T1 $\to$ T1ce are compared by peak signal-to-noise ratio (PSNR), normalized mean squared error (NMSE), and structural similarity index (SSIM)~\cite{yi2019generative}. From Table~\ref{BraTs23}, our method excels in two tasks. Remarkably, with $S^{2}DC$ pre-training, we gain 5.41\% and 0.32\% improvements in SSIM  separately. We notice that other methods fail to get improvement in the task T1 $\to$ T2 compared to From Scratch. That may be due to the deep domain gap between the two sequences, while pre-trained models provide biased initializations. In contrast, $S^{2}DC$ achieves improvement in NMSE and SSIM, showing better ability in this task.

\textbf{Reconstruction task.} We evaluate the capabilities of reconstruction in different methods on the UDPET dataset. Following the previous work~\cite{cui2024prior}, whole-body data at DFR=50 (\textit{i.e.}, 1/50 doses) were selected to explore the reconstruction ability of all methods. This task has been extensively studied, yet our method, $S^{2}DC$, still achieves superior PSNR results.

\textbf{Overall comparisons on 10 downstream datasets.} The results demonstrate that $S^{2}DC$ performs effectively across 10 datasets and 4 tasks. Compared to training from scratch, which achieves an average score of 77.93\%, $S^{2}DC$ pre-training delivers a 3.5\% improvement, reaching 81.43\%. Additionally, $S^{2}DC$ outperforms the second-best SSL method (VoCo~\cite{wu2024voco}, average score 80.65\%) for all tasks, with an average gain of 0.78\%. These findings indicate that $S^{2}DC$ is both reliable and adaptable across various downstream tasks and modalities.

\subsection{Ablation study}
\label{sec:ablationstudy}
\begin{figure}[!htbp]
  \centering
  \includegraphics[width=0.95\columnwidth]{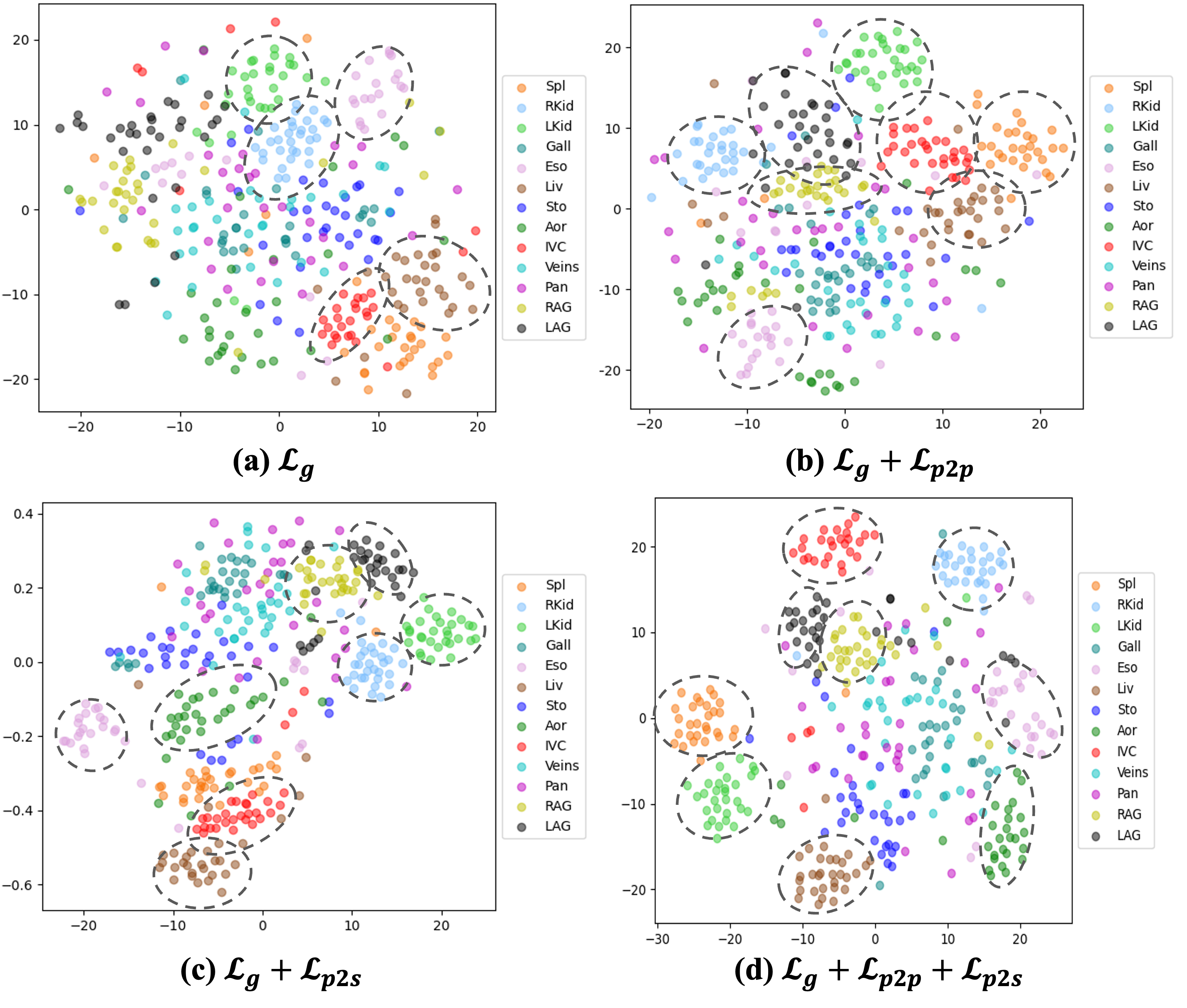}
  \caption{The t-SNE feature visualization of different losses on 13 organs on the BTCV dataset.}
  \label{fig:loss_tsne}
  \vskip -0.1in
\end{figure}

\textbf{Influence of different constraints.}
We use contrastive learning loss $\mathcal{L}_g$ in ~\cref{preliminary} as our baseline. The impact of different constraints in pre-training is presented in \cref{ablation_loss}. The results indicate that combining patch-to-patch loss $\mathcal{L}_{p2p}$ and patch-to-structure loss $\mathcal{L}_{p2s}$ yields SOTA performance compared to~\cref{BTCV}. Meanwhile, we visualize the t-SNE clustering results of 13 organs under different constraints, as shown in ~\cref{fig:loss_tsne}. The $\mathcal{L}_{p2p}$ constraint leads to a sparser inter-structure distribution, while the $\mathcal{L}_{p2s}$ constraint results in more compact intra-structure clusters. Combining $\mathcal{L}_{p2p}$ and $\mathcal{L}_{p2s}$ achieves the best overall clustering performance.


\begin{table}[!htbp]
\centering
\resizebox{0.85\columnwidth}{!}{%
\begin{tabular}{ccccc}
\hline
\multirow{2}{*}{} & \multirow{2}{*}{} & \multirow{2}{*}{} & BTCV & AUTOPET \\ \cline{4-5} 
\textbf{Baseline($\mathcal{L}_g$)} & \textbf{+$\mathcal{L}_{p2p}$} & \textbf{+$\mathcal{L}_{p2s}$} & \multicolumn{2}{c}{DICE(\%)} \\ \hline
 $\bullet$ &  &  & 83.43  & 45.72  \\
 $\bullet$ &  $\bullet$ &  & 83.98 & 45.85 \\
$\bullet$ &  & $\bullet$ & 84.02 & 46.23 \\
$\bullet$ & $\bullet$ & $\bullet$ & 84.14 & 46.47 \\ \hline
\end{tabular}%
}
\vskip -0.05in
\caption{The ablation results of different constraints.} \label{ablation_loss}
\vskip -0.15in
\end{table}



\begin{figure}[!htbp]
  \centering
  \includegraphics[width=0.95\columnwidth]{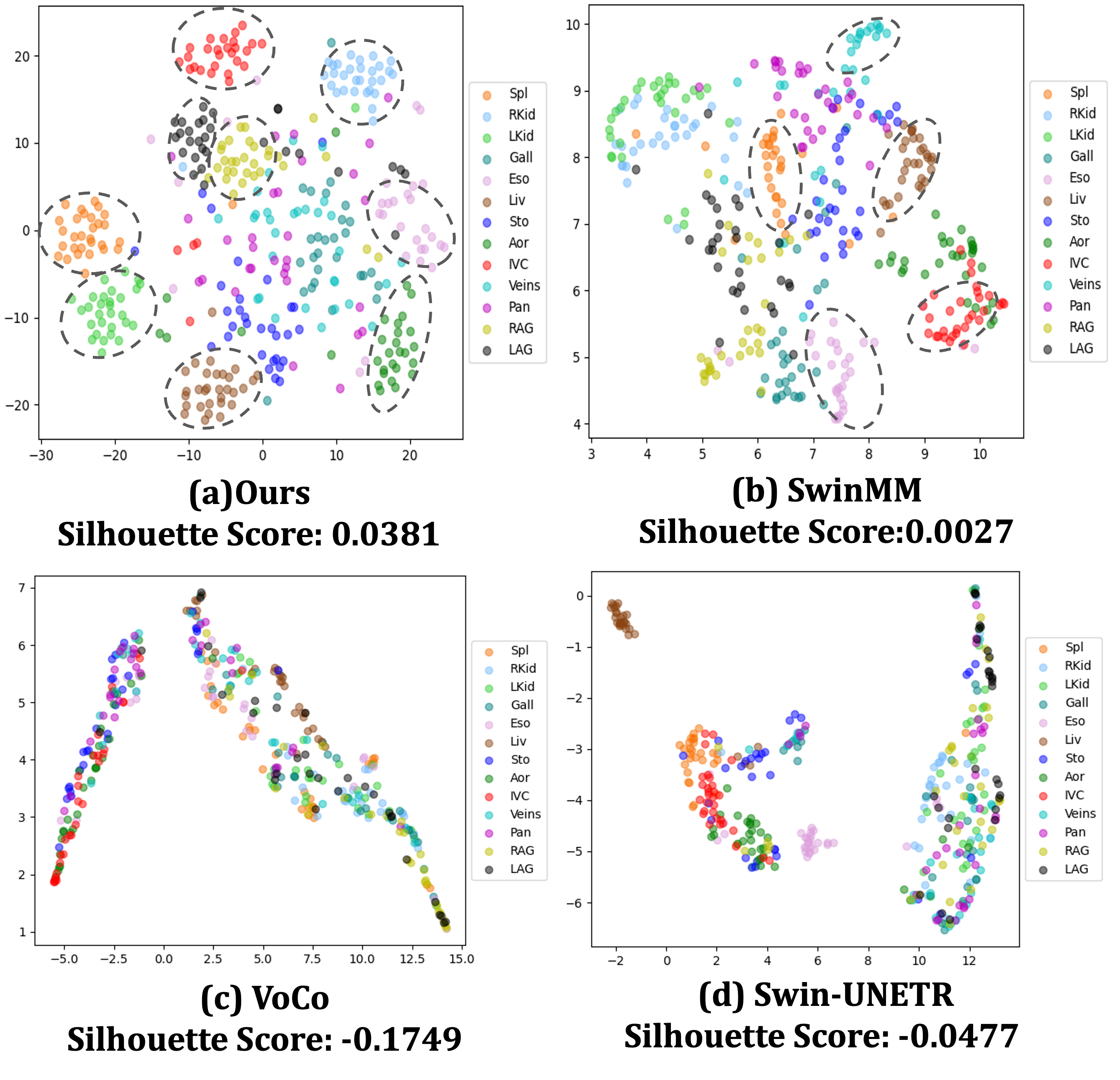}
  \caption{The t-SNE visualization of features from 13 organs on the BTCV dataset and the silhouette scores of four methods.}
  \label{fig:tsne}
  \vskip -0.1in
\end{figure}
\begin{figure}[!htbp]
  \centering
  \includegraphics[width=0.93\columnwidth]{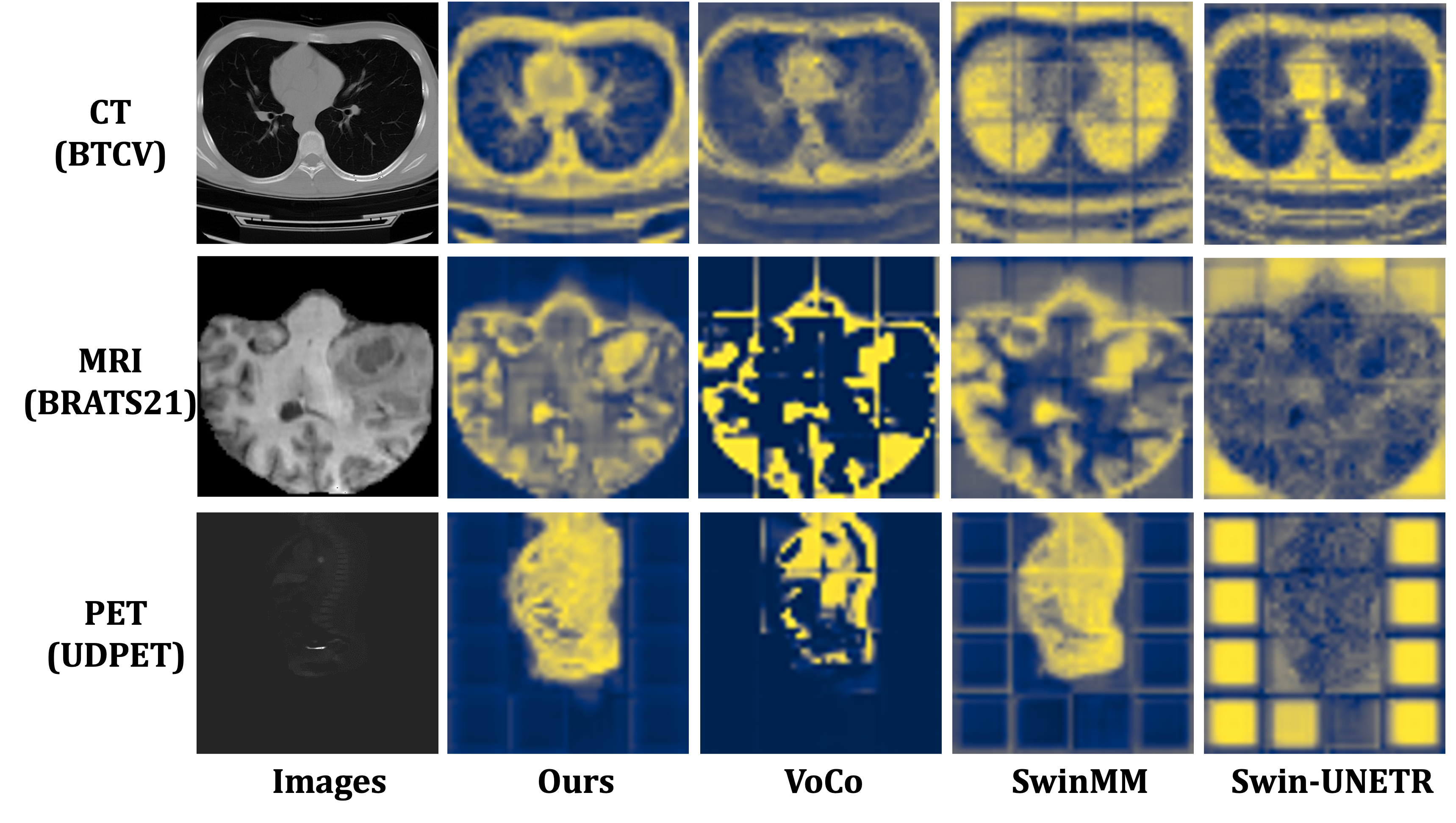}
  \vskip -0.1in
  \caption{We visualize the first principal components after applying PCA to token features~\cite{caron2021emerging} within volumes of CT, MRI, and PET. The figure displays the central slices of feature maps.}
  \label{fig:feature_map}
  \vskip -0.15in
\end{figure}

\noindent\textbf{Intra-structure consistency and inter-structure discrepancy.}
We visualize the t-SNE of features extracted by four medical pre-trained models and calculate their silhouette scores~\cite{rousseeuw1987silhouettes}, which evaluate the quality of clustering by measuring how similar a data point is to its cluster (consistency) compared to other clusters (discrepancy). ~\cref{fig:tsne} shows the clustering capabilities of 13 organs on the BTCV dataset. $S^{2}DC$ shows excellent performance on organ discrimination and consistency, which gets the highest silhouette scores. As shown in ~\cref{BTCV}, most organs that cluster well also exhibit strong segmentation performance, highlighting the importance of structure-specific consistency and discrepancy. Additionally, we visualize the feature maps of 4 SSL models on 3 modalities in \cref{fig:feature_map}. $S^{2}DC$ performs semantic consistency in the background, tissues (CT), functional regions (MRI), and metabolic regions (PET).

\begin{table}[!htbp]
\centering
\resizebox{0.7\columnwidth}{!}{
\begin{tabular}{lccc}
\hline
 & \multicolumn{3}{c}{\textbf{Accuracy(\%)}} \\ \cline{2-4} 
\multirow{-2}{*}{\textbf{Method}} & \textbf{10\%} & \textbf{50\%} & \textbf{100\%} \\ \hline
Swin-UNETR~\cite{he2023swinunetr} & 77.15 & 92.28 & 94.15 \\
SwinMM~\cite{wang2023swinmm} & 87.87 & 93.50 & 94.80 \\
VoCo~\cite{wu2024voco} & 86.73 & 92.43 & 94.60 \\
\textbf{$S^{2}DC$} & \color[HTML]{9A0000}{ \textbf{88.29}} & \color[HTML]{9A0000}{\textbf{93.89}} & \color[HTML]{9A0000}{\textbf{95.34}} \\ \hline
\end{tabular}
}
\vskip -0.05in
\caption{Experiment results on CC-CCII with various ratios of the training data. 10\%, 50\%, and 100\% represent ratios. \textcolor[rgb]{0.25, 0.5, 0.75}{\textbf{Best}} results are highlighted. }\label{varying_ratios}
\vskip -0.1in
\end{table}
\noindent\textbf{Varying ratios of the training dataset.}
We evaluate the performances of models that were trained with varying ratios of CC-CCII. From \cref{varying_ratios}, $S^{2}DC$ outperforms other methods under different ratios of training data. Notably, $S^{2}DC$ achieves 88.29\% accuracy using only 10\% of the data, which is effective with limited dataset sizes.

\begin{figure}[!htbp]
  \centering
  \includegraphics[width=0.75\columnwidth]{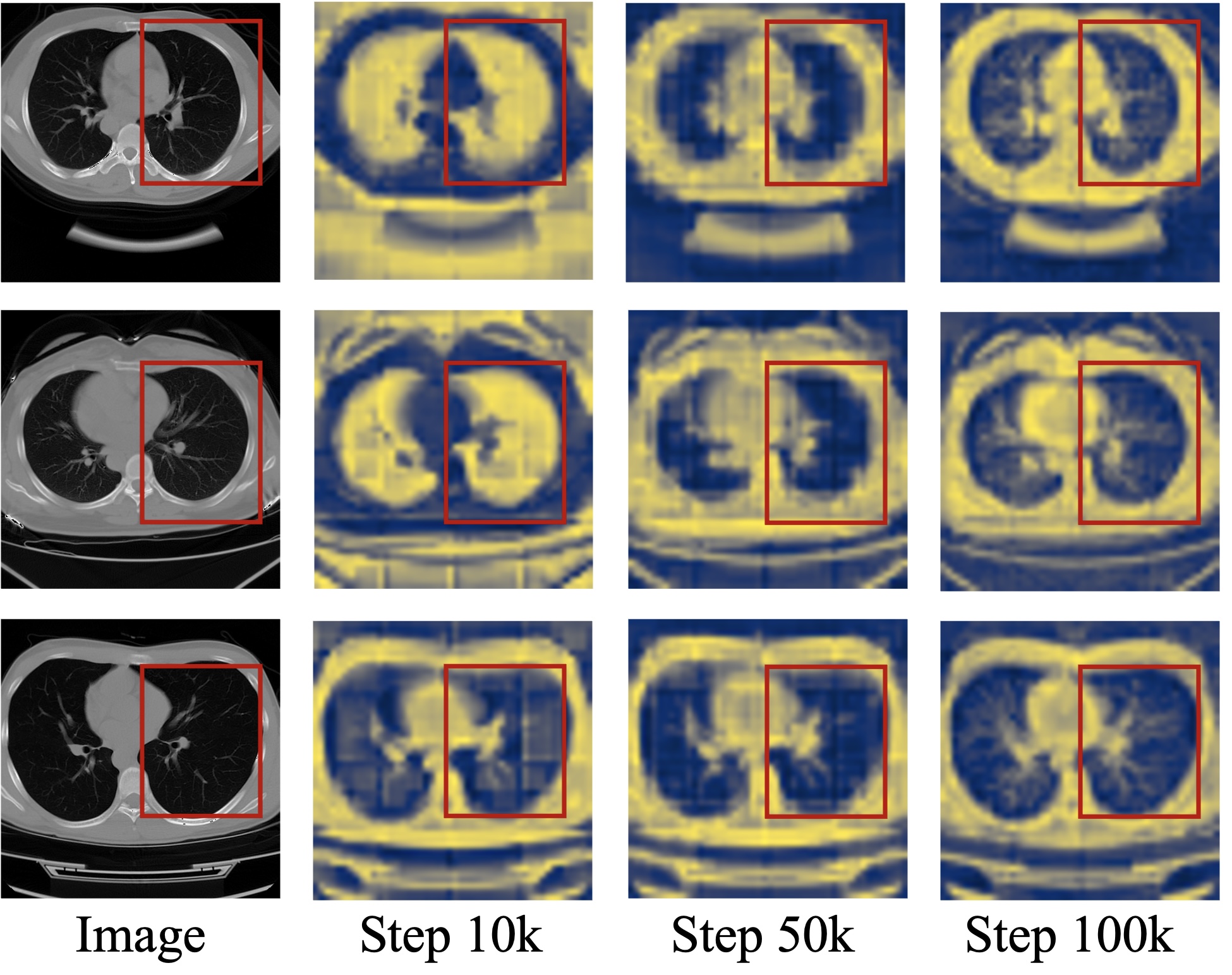}
  \caption{The evolution of patch-to-structure correspondences throughout training.}
  \label{fig:evolution}
  \vskip -0.1in
\end{figure}
\noindent\textbf{Patch-to-structure correspondences evolution.} We visualize the trend of feature maps at different training steps. As shown in ~\cref{fig:evolution}, feature consistency improves as the training progresses. Additionally, larger structures (e.g., the heart) are learned more easily than smaller ones like bronchioles, highlighting the challenge of learning small structures.

\section{Conclusion}
We propose $S^{2}DC$, a medical image SSL framework that enhances structure-aware semantic consistency and discrepancy by patch-to-patch and patch-to-structure steps. Experimental results and visualizations demonstrate $S^{2}DC$'s superior performance. Future exploration could include larger pre-training datasets or model generalization across modalities to support broader applications.

\section{Acknowledgments}
This paper was supported by the National Natural Science Foundation of China (Grant No.82394432 and 92249302), the Shanghai Municipal Science and Technology Major Project (Grant No.2023SHZDZX02). The computations in this research were performed using the CFFF platform of Fudan University.
{
    \small
    \bibliographystyle{ieeenat_fullname}
    \bibliography{main}
}
\end{document}